\documentclass{article}
\usepackage{ijcai24}

\usepackage{times}
\usepackage{soul}
\usepackage{url}
\usepackage[hidelinks]{hyperref}
\usepackage[utf8]{inputenc}
\usepackage[small]{caption}
\usepackage{graphicx}
\usepackage{amsmath}
\usepackage{amsthm}
\usepackage{booktabs}
\usepackage{algorithm}
\usepackage{algorithmic}
\usepackage[switch]{lineno}

\usepackage{natbib}
\usepackage{amsmath}
\usepackage{amssymb}
\usepackage[inkscapelatex=false]{svg}
\usepackage{multirow}
\usepackage{multicol}
\usepackage{booktabs} 
\usepackage{bm}
\usepackage{makecell}
\usepackage{color}
\usepackage{cleveref}
\usepackage{lineno}

\title{A Unified Label-Aware Contrastive Learning Framework for Few-Shot Named Entity Recognition}

\author{Haojie Zhang
\and
Yimeng Zhuang\\
\affiliations
Samsung R\&D Institute China-Beijing\\
\emails
tayee.chang@gmail.com,
ym.zhuang@hotmail.com
}

\begin{document}

\maketitle

\begin{abstract}
Few-shot Named Entity Recognition (NER) aims to extract named entities using only a limited number of labeled examples. Existing contrastive learning methods often suffer from insufficient distinguishability in context vector representation because they either solely rely on label semantics or completely disregard them. To tackle this issue, we propose a unified label-aware token-level contrastive learning
framework. Our approach enriches the context by utilizing label semantics as suffix prompts. Additionally, it simultaneously optimizes context-context and context-label contrastive learning objectives to enhance generalized discriminative contextual representations. Extensive experiments on various traditional test domains (OntoNotes, CoNLL’03, WNUT’17, GUM, I2B2) and the large-scale
few-shot NER dataset (FEW-NERD) demonstrate the effectiveness of our approach. It outperforms prior state-of-the-art models by a significant margin, achieving an average absolute
gain of 7\% in micro F1 scores across most scenarios. Further analysis reveals that our model benefits from its powerful transfer capability and improved contextual representations \footnote{The source code is available at  
 \url{https://github.com/TayeeChang/Unified-Framework-for-FS-NER}}.
\end{abstract}

\section{Introduction}
Named Entity Recognition (NER) is a fundamental task in Natural Language Processing (NLP) that involves extracting entity spans from input sentences and assigning them the correct entity types, such as location, person, and organization \cite{sang2003introduction}. This information is crucial for downstream tasks like question answering, information retrieval, knowledge graph \cite{he2021construction}, machine translation, etc.  However, the process of labeling data is time-consuming and labor-intensive \cite{huang2020few, cui2021template}. Oftentimes, only a few labeled examples are available, and these examples may come from a different domain, posing a significant challenge due to data sparsity \cite{huang2020few, ding2021few}. Consequently, there is growing attention on how to effectively adapt models to handle unseen data in such scenarios.

Recently, there have been two emerging trends in existing methods. One direction revolves around prompt technology. Prompt-based methods focus on bridging the gap between pre-training and fine-tuning by incorporating suitable prompts. These methods exhibit remarkable performance in the few-shot domain, positioning them as promising candidates for both few-shot and zero-shot learning scenarios \cite{ding2021few, chen2022knowprompt}. However, prompt-based methods must carefully design templates \cite{cui2021template, ma2022label}. 

Metric learning is an innovative technology that involves calculating the distance metric between a query token and the reference, and subsequently assigning the corresponding entity type to the query token \cite{hou2020few, yang2020simple}.
Recent advancements in this field have leveraged contrastive learning techniques \cite{das2021container}, leading to state-of-the-art performance. However, it is important to acknowledge that metric learning methods still have limitations. One notable limitation is their tendency to overlook the implicit semantic information underlying entity types, which leads to insufficient exploration of the relationships between different entity types. As a result, the contextual representations may not be sufficiently discriminative.

\begin{figure*}[t]
    \centering
    \includegraphics[scale=0.5]{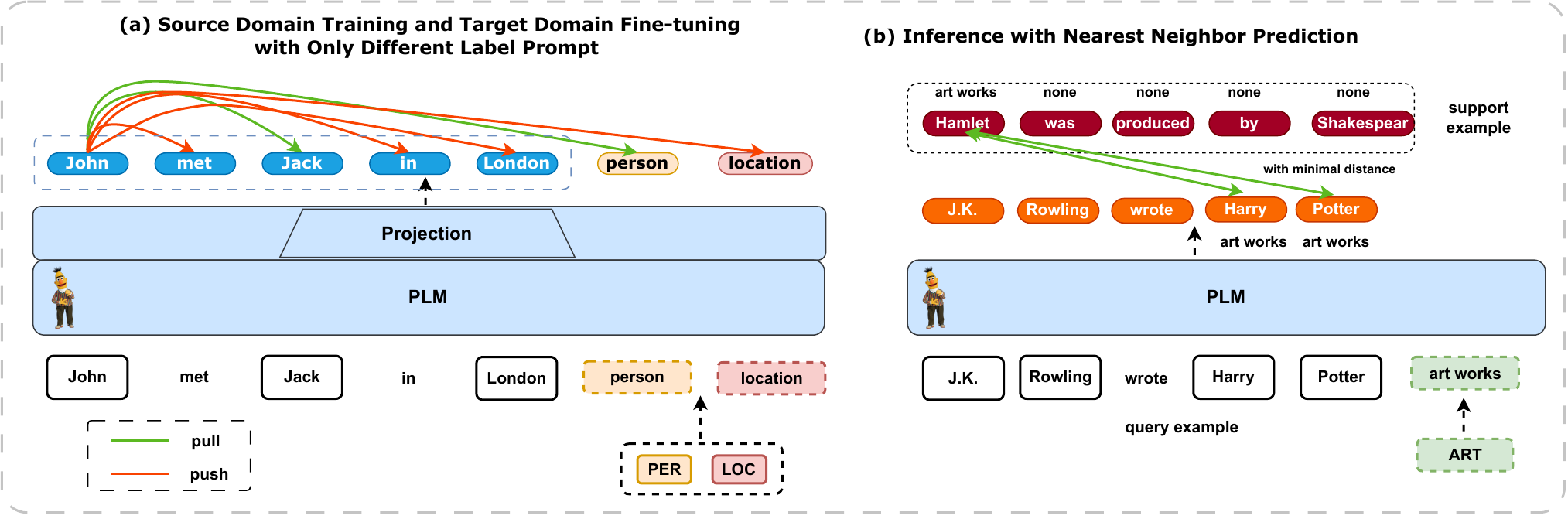}
    \caption{An overview of the architecture of our proposed model. (a) During the training and fine-tuning process in the source domain, the fine-tuning follows a similar approach as training, but with a different label prompt. Utilizing contrastive learning, tokens belonging to the same entity types are attracted toward each other, while tokens representing different entity types are pushed apart. This encourages the model to learn a more distinct and effective representation of entity-specific information. The contrastive learning includes two aspects: context-context and context-label. (b) Inference process with nearest neighbor prediction. Similarity scores between query tokens and support tokens will be calculated according to the distance metric.}
    \label{fig:system_scheme}
\end{figure*}

To tackle these challenges, we propose a label-aware contrastive learning approach that aims to enhance the discriminative power of contextual representations. To enhance the context, we convert entity types into natural language forms and use them as suffix prompts. We also employ both context-to-context and context-to-label contrastive learning, leveraging the connection between context and its corresponding label for improved learning.
By expanding the context using natural language forms and integrating context-to-label contrastive learning, our approach achieves more robust and discriminative contextual representations. As a result, our model demonstrates significant improvements in handling few-shot NER scenarios and addressing data sparsity issues.

Our approach builds upon the contrastive learning method proposed by \citet{das2021container} to learn improved semantic space representations. However, we take three additional steps to address the aforementioned issues. In summary, our contributions can be outlined as follows:

\begin{itemize}
    \item To thoroughly extract the implicit semantic information associated with entity types, we employ label information as a suffix prompt. In particular, we transform the abbreviated form of the entity type into its corresponding natural language description and append it to the original input sentence. This approach allows us to effectively capture the complete meaning conveyed by the entity types.
    \item In addition to constructing context-to-context contrastive learning, we also incorporate context-to-label contrastive learning into our approach. This expansion enables us to leverage the relationship between the context and its associated label, further enhancing the learning process.
    \item We have enhanced the original contrastive loss function at a low cost, which can be seen as an advantageous benefit without any additional expense.

\end{itemize}

\section{Related Work}
\paragraph{Meta Learning} Originating from computer vision, meta learning aims to learn how to learn across tasks \cite{vanschoren2018meta}. Matching Network \cite{vinyals2016matching} employing ideas from the metric learning, was first proposed to adapt to the few-shot scenario. After that, many metric learning based methods are proposed. Prototypical Network \cite{snell2017prototypical} was proposed to aim to learn the prototype representations by computing the average representations of unseen entity types given a support set. \citet{yang2020simple} proposed NNShot and StructShot, which employ the nearest neighbor prediction and Viterbi decoding to inference. Recently, metric learning based methods have become more and more popular in NLP tasks such as text classification \cite{bao2019few, geng2019induction}, named entity recognition \cite{ding2021few, yang2020simple, das2021container, huang2022copner} and machine learning \cite{gu2018meta}.

\paragraph{Prompt Technology} Recently, prompting has become a new paradigm in NLP \cite{liu2023pre, brown2020language}. Many prompt based methods are proposed for varied NLP tasks and show great potentiality in few-shot NER. TemplateNER \cite{cui2021template} utilizes prompt for the entity spans and scores these entity candidates. DemonstrationNER \cite{lee2021good} introduces task demonstrations for in-context learning. Similar to TemplateNER, EntLM \cite{ma2021template} converts sequence labeling task to pre-training LM task, which abandons the prompt but needs to introduce label words relying on a specific domain. \citet{ma2022label} proposed that utilizing label semantics can improve the effect of few-shot NER models especially in the 1-shot scenarios. However, it completely depends on label information to learn the label representations. \citet{huang2022copner} inherits the idea of \citet{ma2022label} by aligning contextual token representations with label representations in the same encoder. But the alignment is limited as it doesn't consider that the contextual tokens include more alignment information. 

\paragraph{Few-shot NER} It aims to adapt the model to unseen entity types. Existing few-shot NER methods present two trends, metric learning based and prompt based. With the great success of prompt technology, many prompt based methods are proposed to apply in the few-shot scenario \cite{cui2021template, lee2021good, ma2021template}. However, these methods require an elaborate prompt design. Recently, label semantics have been proven to be effective for few-shot NER. \citet{ma2022label, huang2022copner} introduce label semantics to learn the label representations by aligning the contextual token representations and label representations. However, both of them 
ignore those contextual token representations that include more alignment information. As a result, as shown in \cite{ma2022label}, with the number of shots increasing, such simple alignment brings marginal effects. 
Metric learning based methods are more popular and gradually become mainstream in recent years \cite{snell2017prototypical, yang2020simple, hou2020few}. In order to better simulate the reality situation, \citet{ding2021few} proposed a new dataset FEW-NERD designed for the few-shot scenario. Inspired by the feature extractors and nearest neighbor inference \cite{wang2019simpleshot}, learning representations and employing nearest neighbor inference seems to have become a standard configuration. As a result, the key point of metric learning based methods is to learn more generalized representations of context tokens. \cite{das2021container} utilizes contrastive learning to learn to distinguish between different tokens. Introducing Gaussian Embedding proves to be more effective in adapting to different domains and achieving current SOTA results in few-shot NER tasks.

\section{Problem Formulation}
NER is generally considered as a sequence labeling task \cite{sang2003introduction, chiu2016named}. For the input sentence $x = \{x_{1},x_{2},...,x_{n}\}$, NER aims to assign a certain label $y_{i}$ for each token $x_{i}$. Generally, $y_{i}  \in \mathcal{C}$, $\mathcal{C}$ refers to a specific label set like \{\texttt{person}, \texttt{location}, \texttt{organization}, \texttt{other}\}. Here, \texttt{other} means that a token doesn't belong to any entity type. 

\subsection{Few-Shot NER}
It's pretty easy to achieve state-of-the-art performance by employing the pre-training language model as the basic encoder when the data resource $\mathcal{D}_{H}$ is large. But when we face the case of low resource data $\mathcal{D}_{L}$, which means only a few examples can be obtained, it poses a new challenge.

Formally, for few-shot NER, the source domain label set is $\mathcal{C}_{H}$ and the target domain label set is $\mathcal{C}_{L}$, $\mathcal{C}_{L} \cap \mathcal{C}_{H} = \emptyset$. Since the target label set is unseen for the source label set, it is very challenging for the model to predict the right label sequence. Thus, it requires the model to have a great transfer capability. Different the high resource data set scenario, for the training and evaluation, we follow the $N$-way $K$-shot setting \cite{ding2021few}, which means the label set size is $N$, and for each label, there are $K$ examples.

\subsection{Evaluation Protocols}
Typically, there are two main evaluation protocols for few-shot NER tasks. One is the episode evaluation. For each episode or task, a support and query pair set is sampled from the high resource data set according to the $N$-way $K$-shot setting \cite{ding2021few}. The model is trained and evaluated on this downsampling data for $T$ turns. The final micro-F1 is calculated over all the episodes. The other one is the low resource evaluation \cite{yang2020simple}. Instead of training and evaluation on each episode, the model is trained on the downsampling $N$-way $K$-shot dataset but evaluated on the whole standard hold-out test set. The operation can also be implemented for $T$ turns and the average value of the micro-F1 is the final metric value. For the traditional datasets (e.g., OntoNotes, CoNLL'03, WNUT'17, GUM, I2B2), we follow the low resource evaluation protocol and for the FEW-NERD dataset, which is explicitly designed for the few-shot NER episode task, we adopt the episode evaluation protocol.

\subsection{Tagging Scheme}
For fair comparison, we adopt the IO tagging scheme for all our experiments, where I-type represents a token that is inside an entity and O-type represents a token that doesn't belong to any entity type.

\section{Method}
We propose a novel approach for few-shot Named Entity Recognition (NER) that integrates metric-based contrastive learning and prompt technology within a unified framework. Our method involves a two-stage training process, starting with training on the source domain and then fine-tuning on the target domain. The key aspect of our proposed method lies in its unified framework, which combines metric-based contrastive learning and prompt technology. Additionally, we have improved upon the original contrastive learning scheme by making slight adjustments to the contrastive loss. This effectively enhances model training and provides a significant boost in performance without any additional cost.

As demonstrated in Figure \ref{fig:system_scheme}, we first train our model in source domain spaces. Next, we finetune our model in the target domain spaces. The fine-tuning process shares the same methods as the training process. Finally, the nearest neighbor prediction is employed to predict the label sequence of the query set in the target domain.

\subsection{Source Domain Training}
Figure \ref{fig:system_scheme} illustrates our unified framework, which combines the original context and label prompt using contrastive learning technology. This technology encompasses two key aspects: 1) context-context contrastive learning and 2) context-label contrastive learning. The goal of contrastive learning is to distinguish the differences between different token representations, including both context-context and context-label relationships.
Label semantics have been proven effective for named entity recognition, particularly in few-shot scenarios \cite{ma2022label}. To incorporate label semantic information, our model employs a suffix prompt. Firstly, we convert abbreviated-form labels to their natural language descriptions to obtain sufficient semantic information. For example, we manually map labels like ⟨PER⟩ to ⟨person⟩, ⟨LOC⟩ to ⟨location⟩, ⟨ORG⟩ to ⟨organization⟩, etc.
The label prompt is constructed by arranging the natural language label sequences separated by a special token [CLS]. Similar to BERT \cite{devlin2018bert}, the [CLS] token represents the entire input sentence and can be used as the representation of the label itself. The context text and label prompt are then concatenated and fed into a pre-trained language model (PLM). A projection layer is applied to project the semantic representations into a different representative space. Finally, we utilize the context-context and context-label contrastive learning losses to train our model effectively.
By combining contrastive learning with label prompts, our unified framework enhances the discriminative power of contextual representations for few-shot NER tasks.

Given a context sequence of $n$ tokens $[x_{c_{1}},x_{c_{2}},...,x_{c_{n}}]$, by label prompt technology, we can get the whole input tokens $[x_{c_{1}},x_{c_{2}},...,x_{c_{n}},x_{l_{1}},x_{l_{2}},...,x_{l_{k}}]$, where $k$ is the whole label number including entity and none-entity and for simplicity, $x_{l_{i}}$ represents $i$-th label token. Here, for simplicity, we omit the word-piece tokenization and the special separative tokens [CLS] and [SEP].

\begin{algorithm}[t]
\caption{Fine-tuning in the target domain}
\label{alg:algorithm}
\textbf{Input}: Support Data $\mathcal{X}_{sup}$, Encoder $\texttt{PLM}$, Projection head $f_\mu$, $f_\Sigma$;\\
\textbf{Output}: $\texttt{PLM}$, $\mu$, $\Sigma$;
\begin{algorithmic}[1]
\STATE $\mathcal{L}_{prev} \in \mathbb{R}_{+}$ (infinity) \\
\STATE $\mathcal{L} = \mathcal{L}_{prev} - 1$
\REPEAT
\STATE $\mathcal{L}_{prev}$ = $\mathcal{L}$
\FOR{ all $(x_{i}, y_{i}) \in \mathcal{X}_{sup}$} 
\STATE Calculate $\ell_{i}$ as in Eq. (\ref{eq:equation}), Eq. (\ref{eq:context-context}), Eq. (\ref{eq:context-label});
\ENDFOR
\STATE Calculate $\mathcal{L}_{\text{context-context}}$ as in Eq. (\ref{eq:context-context-total})
\STATE Calculate $\mathcal{L}_{\text{context-label}}$ as in Eq. (\ref{eq:context-label-total})
\STATE Calculate $\mathcal{L}$ as in Eq. (\ref{eq:mixed equation})
\STATE Update $\texttt{PLM}$, $f_{\mu}$, $f_{\Sigma}$ by gradient descent to reduce $\mathcal{L}$
\UNTIL{$\mathcal{L} > \mathcal{L}_{prev}$}
\end{algorithmic}
\end{algorithm}

After feeding to the \texttt{PLM}, we can get the last hidden layer outputs of \texttt{PLM} as the representations of input tokens.
\begin{equation}
\label{eq:equation}
    [h_{i},h_{2},....h_{m}] = \texttt{PLM}([x_{1},x_{2},...,x_{m}])
\end{equation}

We adopt a projection head for mapping the representations to a different space like \citet{das2021container}.
\begin{equation}
\label{eq:euation1}
\begin{split}
    \mu_{i} = f_{\mu}(h_{i}), \Sigma_{i} &= f_{\Sigma}(h_{i})
\end{split}
\end{equation}

For context-context contrastive learning, we adopt the Jensen–Shannon divergence between valid context tokens in the same batch. JS-divergence is as follows:
\begin{equation}
    d(p, q) = \frac{1}{2}(\mathcal{D}_{\text{KL}}(\mathcal{N}_{p}||\mathcal{N}_{q}) + \mathcal{D}_{\text{KL}}(\mathcal{N}_{q}||\mathcal{N}_{p}))
\end{equation}

where $D_{\text{KL}}$ refers to the Kullback-Leibler divergence.
The contrastive loss ContaiNER \cite{das2021container} employed is as follows:
\begin{equation}
    \ell(p) = -\log\frac{\sum_{(x_{q},y_{q}) \in \chi_{p}} \exp(-d(p,q)) / \left| \chi_{p} \right|}{\sum_{(x_{q},y_{q}) \in \chi,p \neq q} \exp(-d(p, q))}
\end{equation}

Here, we propose a simple variant,
\begin{equation}
\label{eq:context-context}
\resizebox{1\linewidth}{!}{$
    \ell_{\text{out}}(p) = -\frac{1}{\left| \chi_{p} \right|} \sum_{(x_{q},y_{q}) \in \chi_{p}} \log\frac{\exp(-d(p,q))}{\sum_{(x_{q},y_{q}) \in \chi,p \neq q} \exp(-d(p, q))}
$}
\end{equation}

For $p$-th valid token $(x_{p}, y_{p})$, the positive sample set $\chi_{p}$ is composed of tokens with the same label as $(x_{p},y_{p})$ in the batch set $\chi$.

The context-context contrastive loss is as follows:
\begin{equation}
\label{eq:context-context-total}
    \mathcal{L}_{\text{context-context}} = \frac{1}{\left| \chi_{\text{context}} \right|} \sum_{p \in \chi_{\text{context}}} \ell_{out}(p)
\end{equation}

The context-label contrastive loss is as follows:
\begin{equation}
\label{eq:context-label}
    \ell_{\text{context-label}}(p) = -\log\frac{\exp(-d(p,q) / \tau)}{\sum_{k, k \in \mathcal{C}} \exp(-d(p, k) / \tau)}
\end{equation}

where $y_{p} = q$, $q$ is the prompt token and $\mathcal{C}$ is the prompt label set. For the context-label contrastive loss, there is only one positive example of the token $x_{p}$, which is its golden label token in the prompt.
\begin{equation}
\label{eq:context-label-total}
    \mathcal{L}_{\text{context-label}} = \frac{1}{\left| \chi_{\text{context}} \right|} \sum_{p \in \chi_{\text{context}}}\ell_{\text{context-label}}(p)
\end{equation}

As it's very easy to keep tabs on the saturation point to use the temperature coefficient when employing JS-divergence as the distance metric. Hence we set $\tau$ as 1.

The whole training loss will be:
\begin{equation}
\label{eq:mixed equation}
    \mathcal{L} = \alpha \cdot \mathcal{L}_{\text{context-context}} + (1 - \alpha) \cdot \mathcal{L}_{\text{context-label}}
\end{equation}

\begin{table}[t]
    \centering
    \begin{tabular}{lrrr}
    \toprule
    \textbf{Dataset} & \textbf{Domain} & \textbf{\# Sentences} & \textbf{\# Entity} \\
    \midrule
    FEW-NERD & Wikipedia & 188k & 66 \\
    CoNLL'03  & News      & 20.7k  & 4 \\
    GUM      & Wiki      & 3.5k   & 11 \\
    WNUT     & Social    & 5.6k   & 6 \\
    OntoNotes  & Mixed   & 76k    & 18 \\
    I2B2'14   & Medical  & 140K   & 23 \\
    \bottomrule
    \end{tabular}
    \caption{Summary Statistics of Original Datasets}
    \label{tab:1}
\end{table}

\subsection{Fine-tuning in the Target Domain}
After completing the training phase in the source domain, the next step is to fine-tune the model for adaptation to a different target domain. Since the target domain typically has only a limited number of examples, we gather all the examples into a single batch for fine-tuning. The fine-tuning process closely resembles the training process. In the 5-shot setting, we utilize the same training setting as used during training. However, in the 1-shot setting, we find better results by fine-tuning the model using the Euclidean distance metric instead of the Jensen–Shannon divergence. Additionally, in the 1-shot setting, we omit the context-context contrastive loss as it does not contribute significantly to the training process. This suggests that the label semantic information is more crucial in the 1-shot setting. It is worth noting that due to the limited number of examples in the target domain, we implement an early stopping criterion based on contrastive loss to prevent overfitting. Algorithm \ref{alg:algorithm} provides a detailed illustration of the entire process.

\begin{table*}[ht]
    \centering
    \resizebox{\linewidth}{!}{
    \begin{tabular}{lcccccccccc}
    \toprule
    \multirow{2}{*}{\textbf{Model}} & \multicolumn{5}{c}{\textbf{1-shot}} & \multicolumn{5}{c}{\textbf{5-shot}} \\
    \cmidrule(lr){2-6}
    \cmidrule(lr){7-11}
    & \textbf{I2B2} & \textbf{CoNLL} & \textbf{WNUT} & \textbf{GUM} & \textbf{Avg.} & \textbf{I2B2} & \textbf{CoNLL} & \textbf{WNUT} &  \textbf{GUM} & \textbf{Avg.} \\
    \cmidrule(lr){2-6}
    \cmidrule(lr){7-11}
    
    ProtoBERT$^{\dagger}$  & 13.4 $\pm$ 3.0     & 49.9 $\pm$ 8.6   & 17.4 $\pm$ 4.9   & 17.8 $\pm$ 3.5  & 24.6   & 17.9  $\pm$ 1.8  & 61.3 $\pm$ 9.1  & 22.8 $\pm$ 4.5 &  19.5 $\pm$ 3.4 & 30.4\\
    NNShot$^{\dagger}$     & 15.3 $\pm$ 1.6     & 61.2 $\pm$ 10.4   & 22.7 $\pm$ 7.4   & 10.5 $\pm$ 2.9  & 27.4   & 22.0  $\pm$ 1.5  & 74.1 $\pm$ 2.3  & 27.3 $\pm$ 5.4 & 15.9 $\pm$ 1.8 & 34.8\\
    StructShot$^{\dagger}$ & 21.4 $\pm$ 3.8     & 62.4 $\pm$ 10.5   & 24.2 $\pm$ 8.0   & 7.8 $\pm$ 2.1 &  29.0 & 30.3  $\pm$ 2.1  & 74.8 $\pm$ 2.4  & 30.4 $\pm$ 6.5 & 13.3 $\pm$ 1.3 & 37.2 \\
    DecomposedModel \cite{ma2022decomposed} & -  & 46.1 $\pm$ 0.4  &  25.1 $\pm$ 0.2 & 17.5 $\pm$ 1.0  & - &
    - & 58.2 $\pm$ 0.9  &  31.0 $\pm$ 1.3 & 31.4 $\pm$ 0.9  & -  \\
    CONTaiNER \cite{das2021container}  & 16.4 $\pm$ 1.7     & 57.8 $\pm$ 10.7   & 24.2 $\pm$ 2.9   & 17.9 $\pm$ 1.8 & 29.1  & 24.1  $\pm$ 1.9  & 72.8 $\pm$ 2.0  & 27.7 $\pm$ 2.2 & 25.2 $\pm$ 2.7 & 37.3 \\
    COPNER \cite{huang2022copner}     & 34.6 $\pm$ 1.8   & 67.0 $\pm$ 3.8    & 33.8 $\pm$ 2.5   & - & - & 41.1  $\pm$ 1.6  & 74.9 $\pm$ 2.9  & 34.8 $\pm$ 3.1 &  - & - \\ 
    
    \textbf{Our model}  & \textbf{38.4 $\pm$ 5.0}   & \textbf{72.8 $\pm$ 4.3}  & \textbf{35.7 $\pm$ 4.5}  & \textbf{23.9 $\pm$ 7.2} & \textbf{42.7} & \textbf{46.1  $\pm$ 2.1}  & \textbf{77.6 $\pm$ 1.8}  & \textbf{35.8 $\pm$ 2.1} & \textbf{33.7 $\pm$ 4.7}  &  \textbf{48.3} \\ 
    \bottomrule
    \end{tabular}}
    \caption{F1 scores with standard deviations for Domain Transfer Evaluation. $^{\dagger}$ denotes the results reported by \citet{das2021container}. For some baselines, some values are not reported from the original paper and they are marked as -. For a fair comparison, we directly use sampled shots as in \citet{yang2020simple}. The best results are in \textbf{bold}.}
    \label{tab:2}
\end{table*}

\subsection{Inference Process}
During the inference period, we adopt the nearest neighbor prediction to get the predicted label sequence of the query example \cite{wang2019simpleshot}. Similar to CONTaiNER \cite{das2021container}, we only use the \texttt{PLM} hidden layer outputs as the token representations, so the projection heads $f_{\mu}$, $f_{\Sigma}$ are abandoned in this period. 

More specifically, we first obtain the representation of the support token $h_{x^{\prime}}$ from the support set and the representation of a query token $h_{x}$ according to Eq. (\ref{eq:equation}). Then,
for each query token, we calculate its Euclidean distance between its \texttt{PLM} representation and support tokens. Finally, we will assign the query token to the label of its nearest support token.
\begin{equation}
\begin{split}
     y^{\star} = \arg\min_{y \in \mathcal{Y}}{ \min_{x^{\prime} \in \mathcal{S}_{y}} d(h_{x}, h_{x^{\prime}})} \\
     d(h_{x}, h_{x^{\prime}}) = \Vert h_{x} - h_{x^{\prime}} \Vert^{2}
\end{split}
\end{equation}

where $h_{x}$ denotes $x$-th context token in the query set and $h_{x^{\prime}}$ denotes the context token in the support set.

\section{Experiments Setups}
\subsection{Datasets}
Our model is evaluated with different datasets. We utilize 6 different datasets: OntoNotes 5.0 \cite{weischedel2013ontonotes}, WNUT-2017 \cite{derczynski2017results}, CoNLL-2003 \cite{sang2003introduction}, I2B2-2014 \cite{derczynski2017results}, GUM \cite{zeldes2017gum} and FEW-NERD \cite{ding2021few}.
These datasets come from different domains. The first 5 datasets are Mixed, Social, News, Medical, and Wiki. As the entity distribution of these traditional datasets is usually considered uneven, it's not suitable for the few-shot scenario. Hence \citet{ding2021few} proposed a new dataset FEW-NERD. The FEW-NERD dataset is specially developed
for the few-shot NER task, including INTRA and INTER two sub-tasks with a grand total of 66 fine-grained labels across 8 coarse-grained classes. A summary of statistics of these datasets is shown in Table \ref{tab:1}.

\subsection{Evaluation Settings}
We employ two different evaluation settings to assess the performance of our model: the domain transfer setting and the FEW-NERD setting.

\paragraph{Domain Transfer Setting} This setting evaluates the transfer capacity of the model from one domain to another different domain. In the few-shot scenario, the model transfer capacity is critically important. So 
 in this setting, we evaluate our model on different datasets. We consider ONTONOTES 5.0, which is a mixed dataset, as the source domain. We first train our model on the ONTONOTES 5.0 and fine-tune the model on the CoNLL-2003, WNUT-2017, I2B2-2014, and GUM. For each experiment, 5 different support sets are evaluated by the micro-f1 score with mean and standard deviation reported. We show these results in Table \ref{tab:2}. We see that our model outperforms prior SOTAs with a significant margin, even with existing great domain gaps between OntoNotes and target domains (I2B2, WNUT, GUM). When the source domain and target domain (CoNLL) are closer, our model performs considerable performance like supervised learning.

\paragraph{FEW-NERD Setting}  We evaluate our model on the FEW-NERD dataset \cite{ding2021few}. This dataset is the first of its kind, specifically designed for the few-shot NER task, and it is also one of the largest manually annotated datasets available. It comprises 66 fine-grained and 8 coarse-grained entity types, making it more challenging compared to traditional datasets that are down-sampled in few-shot settings. Additionally, the dataset includes two subtasks. 1) \textbf{INTRA}, where fine-grained entity types in train, dev, and test sets belong to different coarse-grained entity types. To get it straight, there is no overlap between train, dev, and test sets in terms of coarse-grained entity types. 2) \textbf{INTER}, where fine-grained entity types in train, dev, and test sets share the same coarse-grained entity types.
Obviously, the FEW-NERD (INTRA) task presents greater challenges compared to the FEW-NERD (INTER) task due to limitations in sharing coarse-grained entity types. Moreover, the FEW-NERD dataset adopts the episode learning approach within the meta-learning framework. Each episode or task is constructed based on the $N$-way $K$-shot setting. Hence, in the FEW-NERD scenario, we strictly follow this episode evaluation protocol. We assess our model's performance on all episodes of the two distinct sub-tasks and calculate the final micro-F1 score across all episodes. As shown in Table \ref{tab:3} and Table \ref{tab:4}, our approach achieves new SOTA results in the leaderboard in both of the sub-tasks.

\subsection{Implementation Details}
We implement our model by \texttt{Tensorflow 2.4}. We leverage \texttt{bert-base-uncased} as our contextual encoder for all experiments. For the optimizer, we use AdamW \cite{loshchilov2017decoupled} with a weight decay rate of 0.01, excluding the Norm layer and the bias terms. We train our mode with a learning rate of $5 \times 10^{-5}$ for both training and fine-tuning processes, batch size of 32, and maximum sequence length of 128. For a fair comparison, we set the Gaussian Embedding dimension fixed to $l = 128$. For the hyperparameter $\alpha$, we perform a grid search and select the best setting with the validation set. For the training process, we only run one epoch, as we find it will fall into the pattern collapse if continuing to train. For the label prompt, we manually convert the labels in abbreviation to the natural language descriptions in all the datasets. We run all experiments on an NVIDIA A6000 GPU.

\begin{table}[t]
    \centering
    \resizebox{\linewidth}{!}{
    \begin{tabular}{lccccc}
    \toprule
    \multirow{2}{*}{\textbf{Model}} & \multicolumn{2}{c}{\textbf{5-way}} & \multicolumn{2}{c}{\textbf{10-way}} & \multirow{2}{*}{\textbf{Avg.}} \\
    \cmidrule(lr){2-3}
    \cmidrule(lr){4-5}
    & \textbf{1$\sim$2 shot}   & \textbf{5$\sim$10 shot}    & \textbf{1$\sim$2 shot}   & \textbf{5$\sim$10 shot} \\
    \cmidrule(lr){2-3}
    \cmidrule(lr){4-5}
    ProtoBERT$^{\dagger}$  & 23.45  & 41.93  & 19.76  & 34.61  & 29.94 \\
    NNShot$^{\dagger}$     & 31.01  & 35.74  & 21.88  & 27.67  & 29.08  \\
    StructShot$^{\dagger}$ & 35.92  & 38.83  & 25.38  & 26.39  & 31.63 \\
    DecomposedModel \cite{ma2022decomposed} & 52.04  & 63.23  & 43.50 & \textbf{56.84}  & 53.90 \\
    CONTaiNER \cite{das2021container}  & 40.43  & 53.70  & 33.84  & 47.49  & 43.87 \\
    COPNER \cite{huang2022copner}     & 53.52  & 58.74  & 44.13  & 51.55  & 51.99 \\
    \textbf{Our model}       & \textbf{56.51}  & \textbf{65.66}  & \textbf{46.42}  & \text{56.82}  & \textbf{56.35} \\
    \bottomrule       
    \end{tabular}}
    \caption{F1 scores on FEW-NERD (INTRA). $^{\dagger}$ denotes the results reported in \citet{das2021container}. The best results are in \textbf{bold}.}
    \label{tab:3}
\end{table}

\begin{table}[t]
    \centering
    \resizebox{\linewidth}{!}{
    \begin{tabular}{lccccc}
    \toprule
    \multirow{2}{*}{\textbf{Model}} & \multicolumn{2}{c}{\textbf{5-way}} & \multicolumn{2}{c}{\textbf{10-way}} & \multirow{2}{*}{\textbf{Avg.}} \\
    \cmidrule(lr){2-3}
    \cmidrule(lr){4-5}
    & \textbf{1$\sim$2 shot}   & \textbf{5$\sim$10 shot}    & \textbf{1$\sim$2 shot}   & \textbf{5$\sim$10 shot} \\
    \cmidrule(lr){2-3}
    \cmidrule(lr){4-5}
    ProtoBERT$^{\dagger}$  & 44.44  & 58.80  & 39.09  & 53.97  & 49.08 \\
    NNShot$^{\dagger}$     & 54.29  & 50.56  & 46.98  & 50.00  & 50.46  \\
    StructShot$^{\dagger}$ & 57.33  & 57.16  & 49.46  & 49.39  & 53.34 \\
    DecomposedModel \cite{ma2022decomposed} & \textbf{68.77}  & 71.62  & 63.26 & 68.32  & 67.99 \\
    CONTaiNER \cite{das2021container}  & 55.95  & 61.83  & 48.35  & 57.12  & 55.81 \\
    COPNER \cite{huang2022copner}    & 65.39  & 67.59  & 59.69  & 62.32  & 63.75 \\
    \textbf{Our model}       & 67.73  & \textbf{71.81}  & \textbf{63.93}  & \textbf{68.65}  & \textbf{68.03} \\
    \bottomrule       
    \end{tabular}}
    \caption{F1 scores on FEW-NERD (INTER). $^{\dagger}$ denotes the results reported in \cite{das2021container}. The best results are in \textbf{bold}.}
    \label{tab:4}
\end{table}

\section{Main Results}
In this session, we present and analyze the comprehensive results of few-shot NER using two evaluation protocols. Additionally, we perform ablation experiments to demonstrate the effectiveness of our proposed method.

\subsection{Baselines}
We compare our model with various state-of-the-art few-shot NER models. These baseline models can be broadly categorized into two groups: prompt-based approaches and metric learning-based approaches.

\begin{itemize}
    \item \textbf{ProtoBERT} is based on the Prototypical NetWork \cite{snell2017prototypical} and employs BERT \cite{devlin2018bert} as the encoder, which aims to learn the prototype of entity types. It belongs to the family of metric learning. 
    \item \textbf{NNShot} is also based on metric learning, which aims to learn the token-level reference \cite{yang2020simple}. It leverages the nearest neighbor prediction for inference. 
    \item \textbf{StructNN} leverages a Viterbi decoder using the source domain to build the abstract transition probability, compared to NNShot \cite{yang2020simple}. 
    \item \textbf{DecomposedModel} is a decomposed meta-learning approach that conducts entity span detection and entity typing respectively \cite{ma2022decomposed}. 
    \item \textbf{CONTaiNER} is based on contrastive learning, which makes them attractive for tokens of the same entity type and repulsive for tokens of the different entity types \cite{das2021container}.
    \item \textbf{COPNER} leverages label information as the prompt and calculates the distance between the input token and the label token 
 \cite{huang2022copner}. It is one of the prompt-based families.
\end{itemize}

\begin{table}[t]
    \centering
    \resizebox{\linewidth}{!}{
    \begin{tabular}{lcc}
    \toprule
    & \textbf{INTRA} & \textbf{INTER} \\
    \midrule
     \text{Ours w/ OCL + w/o context-label CL} & 57.53 & 63.12 \\
     \text{Ours w/ ICL + w/o context-label CL} & 64.35 & 69.62 \\
     \text{Ours w/ ICL + w/ context-label CL} & \textbf{65.66} & \textbf{71.81} \\
    \bottomrule
    \end{tabular}}
    \caption{Ablation study: F1 scores on FEW-NERD 5-way 5$\sim$10-shot are reported.}
    \label{tab:5}
\end{table}

\begin{table*}[ht]
    \centering
    \resizebox{\linewidth}{!}{
    \begin{tabular}{ccc}
    \toprule
    
    \textbf{Our Model} & \textbf{CONTaiNER} & \textbf{Ground Truth} \\
    
    \midrule
    
    \makecell{\textcolor[RGB]{18,220,168}{$\text{skyway committee}_{\text{organization-other}}$} - \\ deals with improvements to the \\
    \textcolor[RGB]{18,220,168}{$\text{st. paul skyway}_{\text{location-road/railway/highway/transit}}$}} & 
    
    \makecell{\textcolor[RGB]{202,12,22}{$\text{skyway committee}_{\text{organization-company}}$} - \\ deals with improvements to the \\
    \textcolor[RGB]{18,220,168}{$\text{st. paul skyway}_{\text{location-road/railway/highway/transit}}$}} &
    
    \makecell{\textcolor[RGB]{18,220,168}{$\text{skyway committee}_{\text{organization-other}}$} - \\ deals with improvements to the \\
    \textcolor[RGB]{18,220,168}{$\text{st. paul skyway}_{\text{location-road/railway/highway/transit}}$}} \\

    \midrule

     \makecell{\textcolor[RGB]{202,12,22}{the} \textcolor[RGB]{18,220,168}{$\text{cook islands football association}_{\text{organization-other}}$} \\ is the governing body of football in the \\
    \textcolor[RGB]{18,220,168}{$\text{cook islands}_{\text{location-island}}$} .} & 
    
   \makecell{\textcolor[RGB]{202,12,22}{$\text{the cook islands football association}_{\text{location-island}}$} \\ is \textcolor[RGB]{202,12,22}{$\text{the governing body of football}_{\text{organization-other}}$} in the \\
    \textcolor[RGB]{18,220,168}{$\text{cook islands}_{\text{location-island}}$} .} & 
    
    \makecell{the \textcolor[RGB]{18,220,168}{$\text{cook islands football association}_{\text{organization-other}}$} \\ is the governing body of football in the \\
    \textcolor[RGB]{18,220,168}{$\text{cook islands}_{\text{location-island}}$} .}  \\

    \midrule
    \makecell{the album is a conceptual work \\ based on their past album \\
    \textcolor[RGB]{18,220,168}{$\text{stream of consciousness}_{\text{art-music}}$} .} & 
    
    \makecell{the album is a conceptual work \\ based on their past album \\ 
    stream of consciousness .} &
    
    \makecell{the album is a conceptual work \\ based on their past album \\
    \textcolor[RGB]{18,220,168}{$\text{stream of consciousness}_{\text{art-music}}$} .} \\

    \bottomrule
    \end{tabular}}
    \caption{Case study based on FEW-NERD dataset. We give three randomly sampled demonstrations where \text{organization-other}, \text{organization-company}, \text{location-road/railway/highway/transit}, \text{location-island},
    \text{art-music} are entity types.
    Here \textcolor[RGB]{18,220,168}{gree} color denotes right results while \textcolor[RGB]{202,12,22}{red} color denotes mistakes.}
    \label{tab:6}
\end{table*}

\subsection{Overall Results}
Table \ref{tab:2}-\ref{tab:4} shows the overall results of our approach along with those reported by the previous SOTAs. Our approach demonstrates significant performance improvements over prior methods, indicating its effectiveness. Specifically, when compared to CONTaiNER, our approach achieves an impressive increase of up to +16.08 F1 scores on the FEW-NERD Evaluation (INTRA, 5-way 1$\sim$2-shot) and +22 F1 scores on the challenging Domain Transfer Evaluation (I2B2). The I2B2 dataset (Medical), with a different domain and its 23 entity types, is more challenging, further emphasizing the robustness of our model in handling challenging scenarios.
Upon closer analysis of the FEW-NERD evaluation results, an interesting phenomenon emerges. In contrast, our approach exhibits a larger improvement in the 1$\sim$2 shot scenario compared to the 5$\sim$10 shot setting when compared to CONTaiNER. Similarly, when compared to COPNER, our method shows a greater margin of improvement in the 5$\sim$10 shot setting compared to the 1$\sim$2 shot setting. Contrastive learning-based methods show effectiveness in utilizing multiple few-shot examples. By optimizing the KL-divergence among diverse Gaussian embeddings, the model can improve contextual representations, particularly when there are fewer sparse few-shot examples, as in the 5$\sim$10 shot scenario.
In contrast, COPNER solely relies on the label semantic information to align contextual token representations with label representations. But the alignment is limited as it doesn’t consider that the contextual tokens include more alignment information.
It shows improvement more prominently in the 1$\sim$2 shot scenarios, as is shown in Table \ref{tab:3} and Table \ref{tab:4}. It reveals that when the few-shot examples are severely sparse, label information can be proper compensation for the sparsity. 

Notably, FEW-NERD (INTER) consistently outperforms FEW-NERD (INTRA), aligning with their respective designs. FEW-NERD (INTER) utilizes coarse-grained entity types shared across train, dev, and test sets, leveraging joint information for enhanced performance. In summary, our method effectively leverages labels to address data sparsity and improves context token representations using more few-shot examples. This demonstrates the effectiveness of our unified framework, combining label-aware prompts and contrastive learning techniques.

\subsection{Ablation Study and Analysis}
To evaluate the effectiveness of key components in our model, we conducted an ablation study using the following settings: 1) Ours w/ ICL + w/ 
context-label CL: it's the final loss we use in our model. ICL refers to
our improved contrastive loss. 2) Ours w/ ICL + w/o context-label CL: we use our improved contrastive loss (ICL) but remove the context-label contrastive loss. 3)
Ours w/ OCL + w/o context-label CL: we use the origin contrastive loss (OCL) from CONTaiNER \cite{das2021container} and remove the context-label contrastive loss.

\paragraph{Effect of our improved contrastive loss (ICL)} According to Table \ref{tab:5}, our improved contrastive loss method outperforms the CONTaiNER method by a significant margin in both the FEW-NERD (INTRA) and FEW-NERD (INTER) datasets. With only minor adjustments to the original contrastive loss, we achieve substantial performance improvements, making it a nearly effortless enhancement.

\paragraph{Effect of our context-label contrastive loss} As depicted in Table \ref{tab:5}, the utilization of the context-label contrastive loss leads to further improvements in F1 scores. This demonstrates the effectiveness of our approach in leveraging context-label information for enhanced performance.

\paragraph{Case Study} Table \ref{tab:6} shows the results of our case study. To assess entity prediction, we randomly sampled three demonstrations from the FEW-NERD dataset (INTRA and INTER). Comparing our model with CONTaiNER, which also utilizes contrastive learning, our method consistently achieves superior predictions for most cases.

We can see that our model performs well in the first case, accurately extracting all entities with correct spans and entity types. In comparison, CONTaiNER correctly predicts the entity span for "skyway committee" but assigns it the wrong entity type of \texttt{organization-company}. In the second case, both our model and CONTaiNER make errors in predicting the entity boundary (including an extra word "the"). However, CONTaiNER further assigns the wrong entity type of \texttt{location-island} to "the Cook Islands Football Association", while our model predicts the correct entity type. Additionally, CONTaiNER incorrectly extracts the non-existent entity "the governing body of football". Overall, our model's performance is notably superior. In the last case, CONTaiNER fails to make any predictions, whereas our model accurately predicts both the entity span and type.

\begin{figure}[t] 
\centering 
\includegraphics[width=0.5\textwidth]{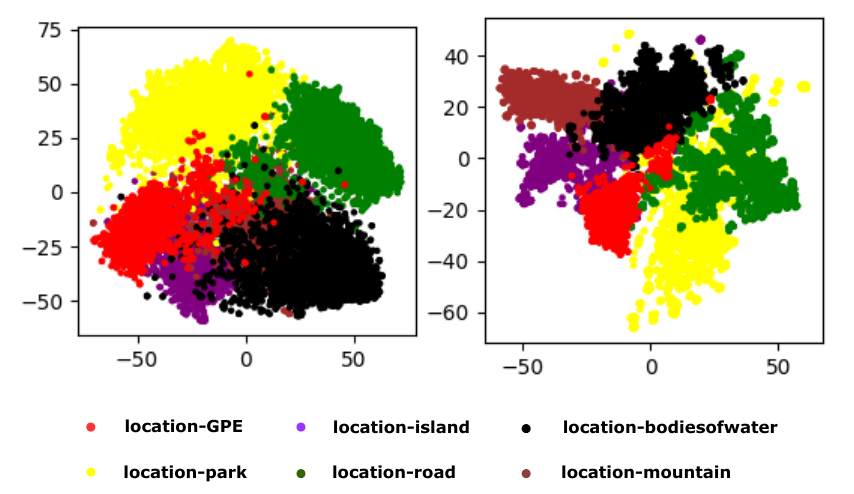} 
\caption{Two-dimensional t-SNE visualizations of the FEW-NERD test set. The token representations are from the sampled 6 fine-grained entity types of \texttt{location} category. The left is for CONTaiNER and the right is for ours.} 
\label{fig:tsne} 
\end{figure}

\paragraph{t-SNE Visualization} We used t-SNE \cite{van2008visualizing} to project BERT's token-level representations onto a 2-dimensional space. Figure \ref{fig:tsne} illustrates the visualization results on the FEW-NERD dataset. In CONTaiNER's visualization (left), there is considerable overlap among different entity classes. Our model (right) effectively preserves a clear distinction between different entity types, highlighting its success in learning a superior metric space specific to each entity type compared to CONTaiNER.

\section{Conclusion}
We propose a unified label-aware contrastive learning framework tailored for few-shot NER tasks. Our model achieves state-of-the-art performance by leveraging label information and optimizing both context-context and context-label contrastive learning objectives. This approach enhances the discriminative power of contextual representations. Extensive experiments on benchmark datasets consistently demonstrate our model's superiority over previous state-of-the-art methods, even in challenging scenarios. In the future, we plan to extend our approach to other token-level classification tasks, such as POS tagging. Additionally, we plan to explore the potential of our model in addressing zero-shot challenges, where the model needs to generalize to unseen label classes.

\bibliographystyle{named}
\bibliography{ijcai24, custom}

\appendix
\section{Additional Implementation Details}
\paragraph{Parameter Setting} We use \texttt{BERT-base-uncased} \footnote{\url{https://github.com/google-research/bert}} as our encoder following Ding et al. (2021). For optimization, we utilize AdamW (Loshchilov and Hutter, 2017) as our optimizer with a fixed learning rate of 5e-5 across all experiments. The embedding dimension is set to $l = 128$, and we maintain a batch size of 32 along with a maximum sequence length of 128. To mitigate overfitting, we apply a dropout rate of 0.1.
During the source domain training, we train the model for only one epoch in all experiments. This measure is implemented to prevent mode collapse, ensuring that the representations of positive and negative examples do not collapse into a single point. Hyperparameter settings are determined through a grid search within the search space outlined in Table \ref{tab:7}.
The source domain training process requires approximately 40 minutes on an NVIDIA A6000 GPU.

\begin{table}[h]
    \centering
    \begin{tabular}{lr}
    \toprule
     Learning rate  & 5e-5 \\
     Batch size & 32 \\
     Max sequence length & 128 \\
     Embedding dimension & 128 \\
     Training epoch & 1 \\
     Optimizer  & AdamW \\
     Alpha   & \{0.8, 0.5, 0.3\}   \\
     
    \bottomrule
    \end{tabular}
    \caption{Experiment setting and hyper-parameter search space in our paper.}
    \label{tab:7}
\end{table}

\section{Fine-tuning Objectives}
As depicted in Table \ref{tab:8}, in the 1-shot scenario, a single example might not adequately capture the distribution information. Therefore, we opt for using the Euclidean distance as the metric instead of the KL-divergence distance. In our Domain Transfer experiments, we observe that optimizing the context-label contrastive loss alone, rather than both losses, yields improved performance in the 1-shot setting. However, in scenarios involving larger data examples like the 5-shot setting and source domain training, we continue to optimize both contrastive losses using the KL-divergence as the distance metric.

\begin{table}[h]
    \centering
    \begin{tabular}{lrr}
    \toprule
      & \textbf{KL-divergence} & \textbf{Euclidean} \\
    \midrule
    1-shot (WNUT)  & 32.83   & 35.72 \\
    1-shot (GUM)   & 18.32   & 23.86 \\
    \bottomrule
    \end{tabular}
    \caption{F1 scores comparison in Domain Transfer Evaluation of 1-shot (WNUT) and 1-shot (GUM) with different fine-tuning objectives.}
    \label{tab:8}
\end{table}

\section{Label Names For Suffix Prompt}

Table \ref{tab:label names} shows the natural language forms of all original labels in all the datasets.

\begin{table}[ht]
    \centering
    \resizebox{\linewidth}{!}{
    \begin{tabular}{lrr}
    \toprule
    \textbf{Dataset} & \textbf{Class} &  \textbf{Natural Language} \\
    \midrule
    \multirow{6}{*}{\textbf{WNUT'17}} & person & person\\
    & location & location \\
    & group & group \\
    & product & product \\
    & creative-work & creative work \\
    & corporation & corporation \\
    \midrule
    \multirow{4}{*}{\textbf{CoNLL'03}} & PER & person \\
    & LOC & location \\
    & ORG & organization \\
    & MISC & miscellaneous \\
    \midrule
    \multirow{9}{*}{\textbf{OntoNotes}} & CARDINAL & cardinal \\
    & DATE & date \\
    & EVENT & event \\
    & FAC & facility \\
    & ORG & organization \\
    & LAW & law \\
    & QUANTITY & quantity \\
    & ORDINAL & number \\
    & $\cdots$  & $\cdots$ \\

    \midrule
    \multirow{9}{*}{\textbf{I2B2}} 
    & \text{DATE} & date \\
    & PATIENT & patient \\
    & DOCTOR & doctor \\
    & HOSPITAL & hospital \\
    & ZIP & zip \\
    & STATE & state \\
    & MEDICALRECORD & record \\
    & HEALTHPLAN & plan \\
    & $\cdots$ & $\cdots$ \\

    \midrule
    \multirow{11}{*}{\textbf{GUM}} 
    & person & person \\
    & place  & place \\
    & organization & organization \\
    & quantity & quantity \\
    & time & time \\
    & event & event \\
    & abstract & abstract \\
    & substance & substance \\
    & object & object \\
    & animal & animal \\
    & plant & plant \\

    \midrule
    \multirow{11}{*}{\textbf{FEW-NERD}} 
    & person-actor & actor \\
    & person-artist/author & artist \\
    & location-island & island \\
    & location-other & location \\
    & organization-education & education \\
    & organization-government/governmentagency & government \\
    & building-airport & airport \\
    & building-other & building \\
    & art-broadcastprogram & broadcastprogram \\
    & other-astronomything & astronomything \\
    & $\cdots$ & $\cdots$\\
    \bottomrule
    \end{tabular}}
    \caption{Natural language form of all label names of all data sets}
    \label{tab:label names}
\end{table}

\end{document}